\documentclass{article}

\usepackage{arxiv}

\usepackage[utf8]{inputenc} 
\usepackage[T1]{fontenc}    
\usepackage{hyperref}       
\usepackage{url}            
\usepackage{booktabs}       
\usepackage{amsfonts}       
\usepackage{nicefrac}       
\usepackage{microtype}      
\usepackage{lipsum}
\usepackage{graphicx}
\graphicspath{ {./images/} }
\usepackage{subcaption}
\usepackage{adjustbox}
\usepackage{multirow}
\usepackage{caption}
\usepackage{wrapfig}
\usepackage{amsmath}
\usepackage{amssymb}
\usepackage{xcolor}

\newtheorem{proposition}{Proposition}

\title{CAST: Closed-form Analytic Semantic Transfer for Zero-Shot Classifier Extension} 

\author{
  William Heyden\thanks{Corresponding author: William Heyden (e-mail: william.heyden@nmbu.no)} ,
  {~~~~~~~~~~} Habib Ullah, 
  {~~~~~~~~~~} M. Salman Siddiqui, 
  {~~~~~~~~~~} Fadi Al Machot\\
  \\
  Faculty of Science and Technology (REALTEK) \\
  Norwegian University of Life Sciences \\ NMBU \\
  1430 Ås, Norway\\
}

\hypersetup{
pdftitle={CAST: Closed-form Analytic Semantic Transfer for Zero-Shot Classifier Extension},
pdfsubject={cs.CV},
pdfauthor={William Heyden, Habib Ullah, M. Salman Siddiqui, Fadi Al Machot},
pdfkeywords={Zero-Shot learning, CLIP, Weigth Injection},
}
\begin{document}
\maketitle

\begin{abstract}
    Large pre-trained models have become foundational components of modern machine learning systems. Yet adapting these models to novel categories typically requires examples from the target distribution.
    In many domains, however, such data are unavailable. Zero-shot learning (ZSL) permits recognition under these limitations through relying on auxiliary semantic information such as textual descriptions.
    We introduce CAST (Closed-form Analytic Semantic Transfer), a training-free, image-free framework for extending a pre-trained classifier to previously unseen classes through weight injection. We provide a theoretical foundation for CAST and derive a finite-sample error decomposition that identifies the \emph{semantic extrapolation residual} $\rho_u$. The residual is a computable, model-agnostic measure and provides a principled criterion for dataset curation and benchmark design. Experiments on standard zero-shot learning benchmarks demonstrate that CAST matches or exceeds existing image-free approaches and approaches the performance of few-shot adaptation methods, while requiring neither iterative optimization nor examples from the target distribution. 
\end{abstract}
\section{Introduction}
Image classification has achieved widespread real-world deployment through large-scale supervised training. However, extending a classifier to new categories demand collecting additional labeled images and retraining the model \cite{pittaras2016comparison, zhou2025revisiting}. In many scientific and industrial applications, however, acquiring data for novel classes is expensive, impractical, or impossible \cite{wang2022whose}.

Zero-shot learning (ZSL) addresses this challenge by transferring knowledge through auxiliary semantic information instead of images \cite{wang2019survey, xian2018zero}. The central premise of ZSL is that knowledge about unseen categories can be transferred through semantic representations such as textual descriptions, attribute vectors, or structured relational information. Recent advances in vision–language models have considerably improved semantic representations, enabling effective transfer to unseen classes \cite{lv2023learning}. However, existing ZSL methods dependent on either image-based optimization or iterative training to learn the semantic transfer, limiting their efficiency and applicability in data-scarce settings.

In this work, we present Closed-form Analytic Semantic Transfer (CAST), an image-free and training-free framework for extending arbitrary pretrained classifiers to previously unseen categories. CAST uses CLIP text embeddings to synthesize classifier weights directly through a closed-form ridge projection. This requirers neither target-domain images nor gradient-based optimization. We further provide a geometric theory showing that zero-shot transfer is governed by semantic extrapolation, leading to the semantic extrapolation residual $\rho_u$, a computable, model-agnostic measure of class-specific transfer difficulty. It provides a principled criterion for semantic data selection and benchmark curation. Experiments demonstrate that CAST achieves competitive zero-shot performance while eliminating the image collection and optimization costs required by existing approaches.

Our contributions are summarized as follows:
\begin{itemize}

\item \textbf{Image-free, training-free classifier extension.}
A closed-form analytical solver for synthesising classifier weights for unseen classes from text descriptions. No target-domain images, no gradient-based optimization.

\item \textbf{A geometric theory of semantic transfer.}
We show the CLIP-text-to-weight-space mapping is the natural linear transformation between two near-optimal semantic representations. By unifying Neural Collapse, cross-modal alignment, and the Linear Representation Hypothesis into one explanation, we derive image-free weight synthesis.

\item \textbf{A data-centric measure of transferability.}
A finite-sample error bound identifies the semantic extrapolation residual $\rho_u$, a computable, model-agnostic predictor of per-class transfer difficulty and a criterion for semantic data selection and benchmark curation.

\item \textbf{Competitive, efficient zero-shot learning.}
CAST matches or exceeds image-free baselines across standard ZSL benchmarks and diverse classifier architectures, while eliminating both image collection and iterative training.

\end{itemize}
\section{Related Work}

\textbf{Embedding-based zero-shot learning} established the canonical ZSL paradigm by learning a shared embedding space between visual features and semantic descriptors~\cite{akata2013label,akata2015evaluation,bucher2016improving}. Early work reduced projection domain shift via bidirectional mappings~\cite{li2017zero}, preserved embedding geometry~\cite{jiang2018learning}, formulated ZSL as cross-domain matching~\cite{zhang2020towards}, and used attention for discriminative representations~\cite{zhu2019semantic}.

\textbf{Generative-based zero-shot learning} instead synthesizes visual features for unseen classes from semantic descriptions, reducing ZSL to a supervised problem trainable on real and generated data alike~\cite{xian2018zero}. Early GAN-based synthesis~\cite{bucher2017generating} was followed by hybrid objectives for quality and stability~\cite{xian2019f,song2018transductive}, class-discriminative generation~\cite{li2019alleviating}, feature diversity~\cite{elhoseiny2021cizsl}, and structured external knowledge~\cite{chen2021knowledge}.

\textbf{Shared limitations.} Both paradigms require seen-class images and task-specific optimization to learn a projection or generative model~\cite{xian2018zero}, so adapting to a new classifier or domain typically means retraining. Embedding-based methods further suffer projection hubness~\cite{dinu2014improving} and seen-class bias~\cite{paul2019semantically,thong2020bias,zhang2024bridging}; generative methods trade these for an expensive two-stage pipeline where feature-synthesis errors propagate to the classifier~\cite{chen2023deconstructed}.

\textbf{Weight imprinting and image-free ZSL.} Weight imprinting assigns classifier weights directly from feature prototypes, which coincide with the optimal classifier geometry under Neural Collapse~\cite{qi2018low,papyan2020prevalence}. Image-free ZSL extends this by synthesizing unseen weights from semantic information alone by a combinations of seen classes~\cite{norouzi2013zero,mensink2014costa} or a learned semantic-to-weight mapping~\cite{christensen2023image}, most recently via LLM-generated class descriptions~\cite{lyu2025expanding} or counterfactual attribute separation~\cite{wang2026counterfactual}. However, a task-specific optimization is required to learn that mapping. CAST derives it analytically instead.

\textbf{Closed-form ZSL.} A separate line of work solves ZSL via closed-form linear estimators. ESZSL~\cite{romeraparedes2015embarrassingly} regularizes a bilinear compatibility function while Shigeto et al.~\cite{shigeto2015ridge} regress semantic labels onto visual prototypes, rather than the reverse, to curb hubness. The seminar work in SAE~\cite{kodirov2017semantic} adds an autoencoder reconstruction constraint solvable via a Sylvester equation. CAST shares this closed-form approach.

\textbf{CLIP and vision-language models.} CLIP~\cite{radford2021learning} aligns images and text in a shared embedding space via large-scale contrastive learning~\cite{dong2025adapting}, later improved through prompt learning and lightweight adaptation~\cite{zhou2022learning,zhou2022conditional,gao2021clip}. CAST instead uses CLIP purely as a semantic text encoder, projecting its embeddings analytically into an arbitrary pre-trained classifier's weight space without modifying the vision model or fine-tuning on additional data.
\section{Preliminaries}
Supervised classifiers are trained on large-scale datasets $\mathcal{D} = \{(x_n, y_n)\}_{n=1}^N$ of image-label pairs $x_n \in \mathcal{X}$ and $y_n \in \mathcal{Y}^s$. Extensive research has demonstrated that trained classifiers admits a canonical decomposition into a feature extractor and a linear classification head~\cite{hild2006feature, rawat2017deep, chen2021review, ciresan2011flexible}:

  \begin{equation}
  \label{eq:cls_decomp}
      f(x) = W^L \circ \phi(x),
  \end{equation}
where $\phi: \mathcal{X} \to \mathbb{R}^d$ maps an input to its penultimate representation and $W^L \in \mathbb{R}^{|\mathcal{Y}^s| \times d}$ is the classification head whose $c$-th row $w_c \in \mathbb{R}^d$ is the learned weight vector for class $c$.

\paragraph{Zero-Shot Learning.}
In Zero-Shot Learning (ZSL) the class space is partitioned into seen classes $\mathcal{Y}^s$, observed during training, and unseen classes $\mathcal{Y}^u$, with $\mathcal{Y}^s \cap \mathcal{Y}^u = \emptyset$. Because the classifier is trained exclusively on $\mathcal{Y}^s$, the head $W^L$ contains no row for any $u \in \mathcal{Y}^u$, making direct classification of unseen classes impossible. ZSL resolves this by associating each class $c \in \mathcal{Y}^s \cup \mathcal{Y}^u$ with a semantic descriptor $a_c$. Under this framing, extending $f$ to $\mathcal{Y}^u$ reduces to estimating one weight vector $\hat{w}_u \in \mathbb{R}^d$ per unseen class.

\paragraph{Neural Collapse.}
Recent work on Neural Collapse (NC)~\cite{papyan2020prevalence} shows classifiers trained to the terminal phase converge to a structured geometric space in the penultimate layer. Letting $\mu_c = \mathbb{E}[\phi(x) \mid y = c]$ denote the class-conditional mean, NC predicts class means and classifier weights co-align into an Equiangular Tight Frame (ETF) and self-duality:
\begin{equation}
      \frac{w_c}{\|w_c\|} = \frac{\mu_c}{\|\mu_c\|} \quad \forall\, c \in \mathcal{Y}^s.
\end{equation}
This means the classifier weight for any class is recoverable from its mean feature vector, which is a well-defined geometric target in $\mathbb{R}^d$.

\paragraph{Weight Imprinting.}
When labeled images of an new class are available, NC directly motivates weight imprinting~\cite{qi2018low}. The synthesised weight is the empirical mean of normalised features,
\begin{equation}
      \hat{w}_u \leftarrow \frac{\bar{\phi}_u}{\|\bar{\phi}_u\|},
      \qquad
      \bar{\phi}_u = \frac{1}{n}\sum_{i=1}^n
          \frac{\phi(x_u^{(i)})}{\|\phi(x_u^{(i)})\|},
\end{equation}
and the new row is appended to the head $\tilde{W}^L = [W^L;\, w_u]$. This represents the few-shot limit. In the zero-shot limit we address requires the same synthesis with \emph{no} images of $\mathcal{Y}^u$, but using only descriptor $a_u$. Given this, we predict the weight vector, $w_u$, the classifier \emph{would have} learned had it seen images of $u$.

\paragraph{CLIP.}
Predicting $w_u$ from $a_u$ alone demands a text encoder whose output space produce a consistent geometric relationship to the classifier's representation space $\mathbb{R}^d$. A Contrastive Language-Image Pre-training (CLIP) model~\cite{radford2021learning} satisfies this by construction. By training two parametric encoders,
  \begin{equation}
  \label{eq:clip_encoding}
      h : \mathcal{X} \to \mathbb{R}^m, \qquad g : \mathcal{A} \to \mathbb{R}^m,
  \end{equation}
via a contrastive objective, CLIP maximise cosine similarity between matched image-text pairs and minimising it for mismatched pairs. This yields a joint embedding space where geometric structure reflects semantic relatedness across modalities. We write $z_c = g(a_c)/\|g(a_c)\| \in \mathbb{S}^{m-1}$ for the unit-normalised text encoding of class $c$.

\section{Method}
With CAST, we imprint weight vectors for unseen classes into a pre-trained classifier, without sample images or re-training. Seen-class weight vectors $w_c$ and their CLIP encoding $z_c \leftarrow g(a_c)$ are predictably structured geometrically, and we aim to learn the map $P: \mathbb{R}^m \rightarrow \mathbb{R}^d$ taking $z_c \mapsto w_c$. Section~\ref{sec:why_bridge_exists} establishes why such a map should exist and generalise to unseen classes; Section~\ref{sec:ridge_objective} derives the closed-form Ridge estimator from seen-class pairs and applies it to synthesize unseen weights (Section~\ref{sec:inference}).

\begin{figure}[t]
    \centering
    \includegraphics[width=0.76\linewidth]{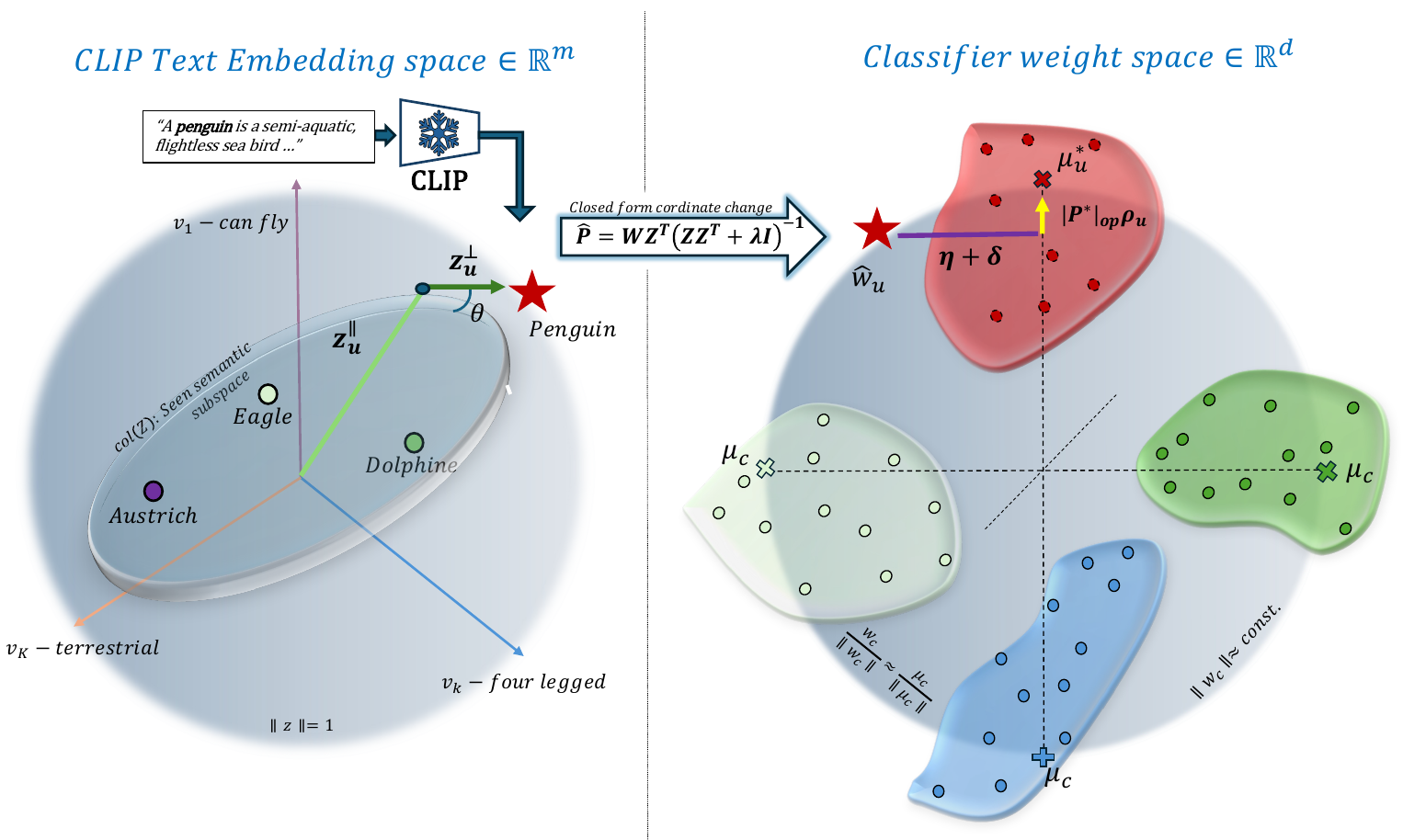}
    \caption{Geometric overview of CAST. \emph{Left:} an unseen class's CLIP text embedding $z_u$ (e.g.\ \emph{Penguin}) decomposes into an in-span component and an out-of-span residual (Eq.~\eqref{eq:semantic_extrapolation_variable}). \emph{Middle:} the Ridge estimator $\hat{P}$ (Eq.~\eqref{eq:ridge_solution}) is the closed-form coordinate change between the two spaces. \emph{Right:} seen-class weights obey a predictable structure; the synthesized $\hat{w}_u$ differs from the true, unobserved $w_u^*$ by the intrinsic gap and the extrapolation penalty (Proposition~\ref{prop:bound}).}
    \label{fig:overview}
\end{figure}

\subsection{Problem Formulation}
\label{sec:problem_formulation}
Our image-free ZSL setting assumes a pre-trained classifier $f$ and a contrastive image-language text encoder $g$, the class $c_u \in \mathcal{Y}^u$ we aim to update our classifier $f$ with, and their associate textual description $a_u$. No image of $c_u$ is available. We stack the seen-class pairs column-wise into the descriptor matrix $Z \in \mathbb{R}^{m \times |\mathcal{Y}^s|}$ and the weight matrix $W \in \mathbb{R}^{d \times |\mathcal{Y}^s|}$ (where \ $W = (W^L)^\top$ from Eq.~\eqref{eq:cls_decomp}), both of which are observed. The goal is to synthesize our target weights $w_u \in \mathbb{R}^d$ for each $u \in \mathcal{Y}^u$.


\subsection{Why the bridge exists}
\label{sec:why_bridge_exists}
We argue why a single linear map $P: \mathbb{R}^m \to \mathbb{R}^d$ from CLIP text embeddings to classifier weights should exist and generalise to unseen classes. The argument rests on three results.

\paragraph{Neural Collapse makes $\mu_c$ the target.} Under NC3, $w_c/\|w_c\| = \mu_c/\|\mu_c\|$: the classifier weight for class $c$ is recoverable from the class-conditional feature mean $\mu_c \leftarrow \phi(x_c)$. Synthesizing $w_c$ from $z_c$ therefore reduces to predicting $\mu_c$ from $z_c$.

\paragraph{CLIP cross-modal alignment gives $P$ on seen classes.} Near-optimal contrastive training provably yields a linear map from CLIP text embeddings to visual class prototypes: building on the transferability analysis of Chen et al.~\cite{chen2024understanding}, there exists a linear map $P$ such that
\begin{equation}
\label{eq:clip_alignment_bound}
    \|Pz_c - \mu_c\| \leq \delta \quad \forall\, c \in \mathcal{Y}^s, \qquad \delta = O(\varepsilon_{\text{CLIP}}^{1/2}),
\end{equation}
where $\varepsilon_{\text{CLIP}}$ is the gap of the CLIP model from the global minimum of its contrastive loss. The contrastive training absorbs the nonlinear alignment of text and image representations. The remaining between text embeddings and classifier weights is a simple linear relationship, consistent with the successful implementations of linear probing on CLIP features \cite{hentschel2022clip}.

\paragraph{The Linear Representation Hypothesis explains why $P$ generalises.} Eq.~\eqref{eq:clip_alignment_bound} only enables $P$ to be accurate for the classes it was estimated on. Generalisation to an unseen class $u$ requires $P$ to act as a stable, concept-level map rather than a seen-class-specific lookup table. Under the LRH~\cite{park2023linear}, a semantic property corresponds to a fixed direction in embedding space rather than a region, the same principle behind embedding arithmetic such as $king - man + woman \approx queen$~\cite{vylomova2016take}. Formally, for each semantic concept $k \in \mathcal{K}$ there exists a direction $v_k \in \mathbb{R}^m$ such that
\begin{equation}
\label{eq:lhr_basic}
    \delta(x) = \sum_{k \in \mathcal{K}(x)} \alpha_k(x) \cdot v_k + \varepsilon(x)
\end{equation}
for a neural encoder $\delta$, where $\mathcal{K}(x)$ is the set of concepts expressed in $x$, $\alpha_k(x)$ is the degree of presence of concept $v_k$, and $\varepsilon$ is noise. Applying Eq.~\eqref{eq:lhr_basic} to the text encoder $g$ and the classifier's feature extractor $\phi$ gives, for each class $c$,
\begin{equation}
  \label{eq:lrh}
  z_c = \!\sum_{k \in \mathcal{K}(c)}\! \alpha_k^c\,v_k^{\text{CLIP}}
        + \varepsilon(a_c),
  \qquad
  \mu_c = \!\sum_{k \in \mathcal{K}(c)}\! \alpha_k^c\,v_k^{\text{vis}}
          + \varepsilon(x).
\end{equation}
A class is described by its concept composition, what changes between modalities is the geometric realisation of those concepts. Combined with Eq.~\eqref{eq:clip_alignment_bound}, this means $P$ acts as a map between concept bases,
\begin{equation}
  \label{eq:lrh_map}
  P\,v_k^{\text{CLIP}} = v_k^{\text{vis}} \quad \forall\,k,
\end{equation}
Any class whose concept composition lies in the span of concepts already expressed by seen classes is covered by the same $P$, whether or not that specific class was observed.

\paragraph{Representation Equivalence is the convergence of these results.} Given two networks $\delta_1: \mathcal{X} \to \mathbb{R}^{d_1}$ and $\delta_2: \mathcal{X} \to \mathbb{R}^{d_2}$ trained to near-optimal representations of the same task, representation equivalence~\cite{lenc2015understanding} states that they are linearly related up to a bounded error,
\begin{equation}
    \mathbb{E}_{x \sim \mathcal{X}}\!\left[\|\delta_2(x) - M\delta_1(x)\|^2\right] \leq \varepsilon,
\end{equation}
for some linear map $M$. CLIP's text encoder $g$ and the classifier's feature extractor $\phi$ are both near-optimal representations of the same visual-semantic concepts. This is the same principle under the linear bound of Eq.~\eqref{eq:clip_alignment_bound} hence, not an empirical coincidence.

\paragraph{Conclusion.} Combining NC3 with Eq.~\eqref{eq:clip_alignment_bound},
\begin{equation}
\label{eq:p_composition}
    w_c \approx \mu_c \approx P z_c,
\end{equation}
and the LRH argument above justifies that the \emph{same} $P$, estimated from seen-class pairs, is the right object to apply unchanged to unseen classes. This is further explored in Section~\ref{sec:ridge_objective}.

\subsection{Ridge Regression and Semantic Regularization ($\lambda$)}
\label{sec:ridge_objective}
A linear map $P$ is identifiable from seen-class pairs $\{(z_c,w_c)\}_{c\in\mathcal{Y}^s}$ by minimizing the seen-class residual $\|PZ-W\|_F^2$. The unregularized solution $\hat{P}_{\text{ols}}=WZ^\top(ZZ^\top)^\dagger$ is ill-conditioned as$Z\in\mathbb{R}^{m\times|\mathcal{Y}^s|}$ has $|\mathcal{Y}^s|\ll m$, and CLIP embeddings concentrate in a low-effective-rank subspace. Therefore, the pseudoinverse amplifies small singular values by $1/\sigma_i$ where the training signal is weakest. We therefore add a Frobenius-norm penalty on $P$,
\begin{equation}
\label{eq:ridge}
  \hat{P} = \arg\min_P\; \|PZ - W\|_F^2 + \lambda\|P\|_F^2,
\end{equation}
whose unique closed-form minimiser
\begin{equation}
\label{eq:ridge_solution}
  \hat{P} = WZ^\top(ZZ^\top + \lambda I)^{-1}
\end{equation}
replaces $1/\sigma_i$ amplification with the shrinkage factor $\sigma_i/(\sigma_i^2+\lambda)$ (full derivation in the supplementary material). Writing $Z=U\Sigma V^\top$, the left singular vectors $u_k$ are, under the LRH, the principal semantic axes of the seen-class concepts (Eq.~\ref{eq:lhr_basic}), and the shrinkage form
\begin{equation}
\label{eq:ridge_svd}
    \hat{P}_\lambda = W V\operatorname{diag}\!\left(\frac{\sigma_k}{\sigma_k^2 + \lambda}\right)U^\top
\end{equation}
assigns each direction $u_k$ a weight $\gamma_k=\sigma_k^2/(\sigma_k^2+\lambda)$. Consistently expressed concepts ($\sigma_k$ large) ensures $\gamma_k\to1$ while sparsely expressed ones ensures $\gamma_k\to0$. Thus $\lambda$ is a control on how much evidence a concept direction needs before being trusted and because CLIP embeddings are $\ell_2$-normalised ($\operatorname{tr}(ZZ^\top)=|\mathcal{Y}^s|$, average non-zero eigenvalue $1$), $\lambda=1$ is a principled choice. In the limit, concept directions entirely absent from the seen-class vocabulary have $\sigma_k=0$ and are mapped to zero by $\hat{P}$. This blind spot is shared by \emph{any} linear estimator trained on seen-class pairs, which we formalise in Section~\ref{sec:semantic_extrapolation_residual} as the \emph{semantic extrapolation residual} $\rho_u$.

\subsection{Inference}
\label{sec:inference}
At test time, synthesizing a weight vector for an unseen class $u$ requires only its textual description $a_u$. The description is encoded through the frozen CLIP text encoder to give $z_u = g(a_u) \in \mathbb{R}^m$, and the projected weight is obtained by a single matrix–vector product:
\begin{equation}
\label{eq:inference}
    \hat{w}_u = \hat{P}\, z_u \in \mathbb{R}^d.
\end{equation}

\paragraph{In the Generalised Zero-shot setting (GZSL),} test images may belong to either seen or unseen classes. Naively concatenating $\hat{W}^u = [\hat{w}_u]_{u \in \mathcal{Y}^u}$ with $W^L$ and taking a joint $\arg\max$ introduces a systematic bias towards seen classes as the rows of $W^L$ are estimated by cross-entropy training, whereas the rows of $\hat{W}^u$ are Ridge projections which inhabit differently scaled geometric structures. We therefore classify with two independent heads. Let
\begin{equation}
\label{eq:gzsl_logits}
    \ell^s = W^L\,\phi(x) \in \mathbb{R}^{|\mathcal{Y}^s|},
    \qquad
    \ell^u = \hat{W}^u\,\phi(x) \in \mathbb{R}^{|\mathcal{Y}^u|}
\end{equation}
be the seen and unseen logit vectors for a test image $x$. A temperature-scaled softmax is applied independently to each head,
\begin{equation}
\label{eq:gzsl_softmax}
    p^s = \operatorname{softmax}\!\left(\ell^s / \tau_s\right),
    \qquad
    p^u = \operatorname{softmax}\!\left(\ell^u / \tau_u\right),
\end{equation}
converting raw scores into within-head confidence distributions that are comparable regardless of weight scale. The image is assigned to the group whose head is more confident, and predicted within that group:
\begin{equation}
\label{eq:gzsl_routing}
    \hat{y} =
    \begin{cases}
        \displaystyle\arg\max_{c \in \mathcal{Y}^s} p^s_c
            & \text{if } \max_c\, p^s_c \;\geq\; \max_c\, p^u_c, \\[6pt]
        \displaystyle\arg\max_{c \in \mathcal{Y}^u} p^u_c
            & \text{otherwise.}
    \end{cases}
\end{equation}
Setting $\tau_s = \tau_u = 1$ recovers a parameter-free baseline; the temperatures provide a post-hoc calibration lever when held-out seen classes are available. No images of class $u$, no gradient computation, and no modification to the backbone are required at any stage.

\paragraph{In Conventional Zero-shot setting (ZSL)} we know that the test image belongs to an unseen class, hence the inference problem reduces to replacing $W^L$ with $\hat{W}^u$ entirely.
\section{Theoretical Analysis}
\label{sec:theoretical_analysis}
The argument chain of Section \ref{sec:why_bridge_exists} justifies why the linear bridge exists. In this section we justify how accurately it can be estimated from a finite set of seen classes and validating each assumption empirically.
Throughout, $\hat{\mu}_c$ denotes the finite-sample estimate of the class-conditional mean $\mu_c = \mathbb{E}[\phi(x) \mid y=c]$; for an unseen class $u$, $\hat{\mu}_u$ is the (never directly observed) estimate that would be obtained from images of $u$, were any available. We work under three assumptions, corresponding to the three results of Section~\ref{sec:why_bridge_exists}: \textbf{(A1)} NC3 holds with rate $\eta$: $\|w_c - \hat{\mu}_c\| \leq \eta$ for all $c$, including unseen $c$~\cite{papyan2020prevalence}; \textbf{(A2)} there exists an ideal map $P^*$ established via CLIP cross-modal alignment (Eq.~\eqref{eq:clip_alignment_bound}) with $\|P^*z_c - \hat{\mu}_c\| \leq \delta$ on seen classes, where $\delta = O(\varepsilon_{\text{CLIP}}^{1/2})$~\cite{chen2024understanding}; \textbf{(A3)} under the LRH (Eq.~\eqref{eq:lrh_map}), the same $P^*$ extrapolates beyond the seen span: $\|P^*z_u - \hat{\mu}_u\| \leq \delta + \|P^*\|_{\operatorname{op}}\,\rho_u$, where $\|P^*\|_{\operatorname{op}} < \infty$ and $\rho_u$ (formalised in Section~\ref{sec:semantic_extrapolation_residual}) measures how far $z_u$ lies outside the span of seen-class embeddings.

\begin{figure}[!h]
    \centering
    \begin{subfigure}{.48\textwidth}
        \centering
        \includegraphics[width=\linewidth]{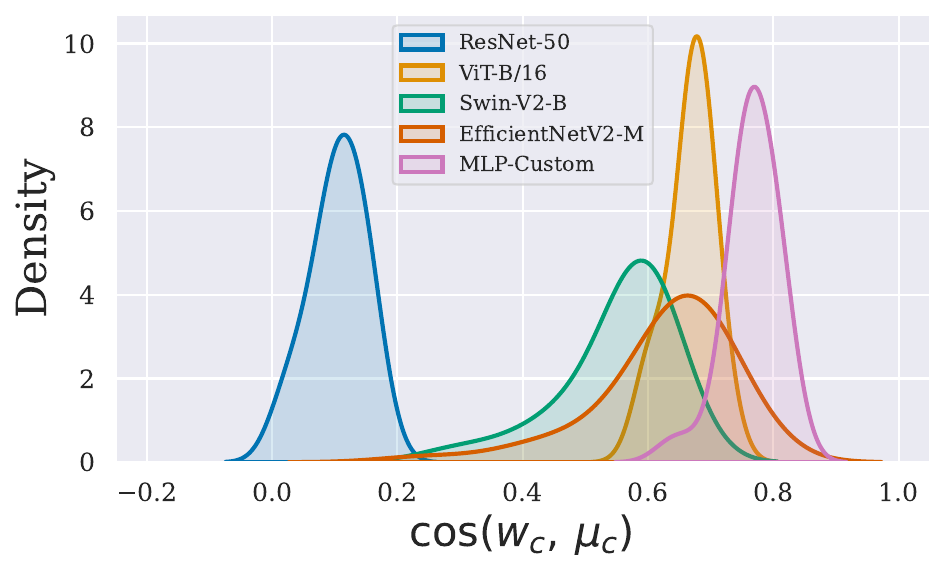}
        \caption{}
        \label{fig:nc3_distribution_awa}
    \end{subfigure}
    \begin{subfigure}{.48\textwidth}
        \centering
        \includegraphics[width=\linewidth]{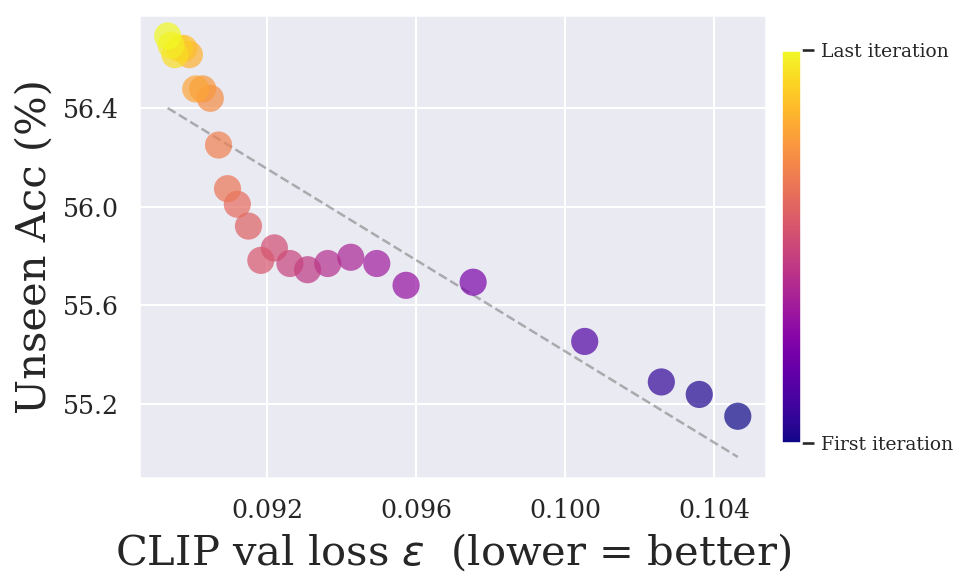}
        \caption{}
        \label{fig:clip_training_awa}
    \end{subfigure}
    \caption{Empirical validation of A1 and A2 on AWA2. (\subref{fig:nc3_distribution_awa})~Cosine alignment $\cos(w_c,\mu_c)$ across five architectures (NC3): architectures satisfying NC3 by design concentrate near $\cos=1$. (\subref{fig:clip_training_awa})~Unseen accuracy vs.\ CLIP validation loss during training, rising monotonically as $\varepsilon\to0$ (A2).}
    \label{fig:nc3_clip_validation}
\end{figure}

In Figure~\ref{fig:nc3_clip_validation} we show that both A1 and A2 are supported empirically. In Figure~\ref{fig:nc3_distribution_awa} we observe that $\cos(w_c,\mu_c)$ shifts towards $1$ for architectures satisfying NC3 by design. However complete NC3 is not necessary for reasonable performance, but rather sets a ceiling on CAST's accuracy. Moderately aligned architectures still work, at the cost of a larger residual. In Figure~\ref{fig:clip_training_awa} its evident that unseen accuracy rises monotonically as CLIP validation loss falls, tracking $\delta(\varepsilon)\to0$. The last assumption is shown in Figure~\ref{fig:rho_vs_acc} and further discussed in Section~\ref{sec:semantic_extrapolation_residual}. 

\subsection{Error Decomposition}
\begin{proposition}[CAST error bound]
\label{prop:bound}
Under the above assumptions, the weight synthesis error satisfies
\begin{equation}
  \label{eq:bound}
  \|\hat{w}_u - w_u^*\|
  \leq O\!\left(
      \underbrace{\eta}_{\text{Classifier gap (NC)}}
    + \underbrace{\delta}_{\text{CLIP gap}}
    + \underbrace{\|P^*\|_{\emph{op}}\,\rho_u}_{\text{Unseen extrapolation}}
  \right)
\end{equation}
\end{proposition}

The three terms follow from the argument chain. By insert $P^*z_u$ and $\hat{\mu}_u$ we can simultaneously pivot the equation:
\begin{equation}
  \label{eq:chain}
  \|\hat{w}_u - w_u^*\|
  \leq
  \underbrace{\|(\hat{P} - P^*)z_u\|}_{(i)}
  +\underbrace{\|P^*z_u - \hat{\mu}_u\|}_{(ii)}
  +\underbrace{\|\hat{\mu}_u - w_u^*\|}_{(iii)}
\end{equation}

It follows that
\begin{align}
  (iii) &\leq \eta
        &&\text{(NC3, A1)} \nonumber\\
  (ii)  &\leq \delta + \|P^*\|_{\text{op}}\,\rho_u
        &&\text{(LRH generalisation to unseen } u\text{, A3)} \nonumber\\
  (i)   &\leq \|P^*\|_{\text{op}}\,\rho_u
        &&\text{($\hat{P} \mapsto z_u^{\perp} = 0$; $P^*$ does not, gap $= \parallel P^*z_u^{\perp} \parallel$)} \nonumber
\end{align}
Summing gives $\|\hat{w}_u - w_u^*\| \leq \eta + \delta + 2\|P^*\|_{\text{op}}\,\rho_u$, i.e.\ Eq.~\eqref{eq:bound} once the constant factor is absorbed into the $O(\cdot)$. Note that this makes it clear that $\rho_u$ binds from two directions (terms (i) and (ii)). The fully explicit derivation, including the residual conditioning term the $O(\cdot)$ absorbs here, is given in the supplementary material.

\subsection{The Semantic Extrapolation Residual}
\label{sec:semantic_extrapolation_residual}
The extrapolation term of Eq.~\eqref{eq:bound} is the most critical for ZSL. This is demonstrated by decompose $z_u = z_u^{\parallel} + z_u^{\perp}$. Here, $z_u^{\parallel} = ZZ^\dagger z_u \in \operatorname{col}(Z)$ is the in-span component the bridge $\hat{P}$ has learned from seen-class data, and $\|z_u^{\perp}\| = \rho_u$ is the out-of-span, untrained direction. For any $v \perp \operatorname{col}(Z)$, $Z^\top v = 0$ implies $\hat{P}v = WZ^\top(ZZ^\top + \lambda I)^{-1}v = 0$: the Ridge estimator is blind to $z_u^{\perp}$.

We define the \emph{semantic extrapolation residual} as this blind component's magnitude,
\begin{equation}
\label{eq:semantic_extrapolation_variable}
  \rho_u = \|(I - ZZ^\dagger) z_u\| = \|z_u^\perp\|.
\end{equation}
Since CLIP embeddings are $\ell_2$-normalised, $\rho_u = \sin(\theta_u)$, the sine of the angle between $z_u$ and its nearest point in $\operatorname{col}(Z)$, so $\rho_u\in[0,1]$: near $0$ for classes lying in the seen-class semantic subspace and recovered by interpolation, approaching $1$ for classes semantically orthogonal to everything seen, where the bridge must extrapolate blindly.

\begin{figure}[t]
    \centering
    \includegraphics[width=0.4\linewidth]{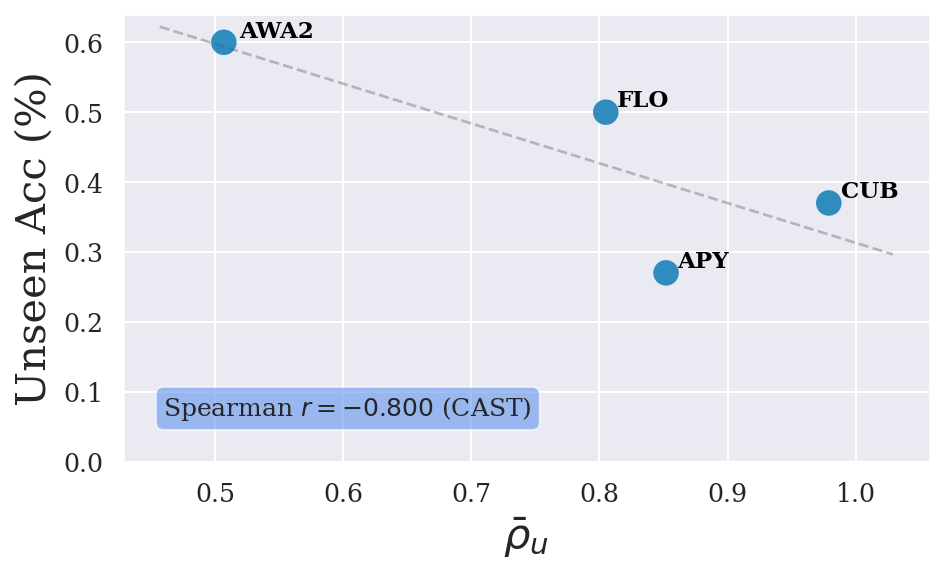}
    \caption{Unseen accuracy vs.\ mean semantic extrapolation residual $\bar{\rho}_u$ across AWA2, CUB, APY, and FLO. Higher $\bar{\rho}_u$ consistently predicts lower accuracy (A3), independently of architecture.}
    \label{fig:rho_vs_acc}
\end{figure}

This is the more critical, class-varying term of the bound (A3), and Figure~\ref{fig:rho_vs_acc} confirms it empirically: mean unseen accuracy versus mean $\bar{\rho}_u$ across all four benchmarks falls on a clear decreasing trend.

\begin{figure}[t]
    \centering
    \includegraphics[width=0.42\linewidth]{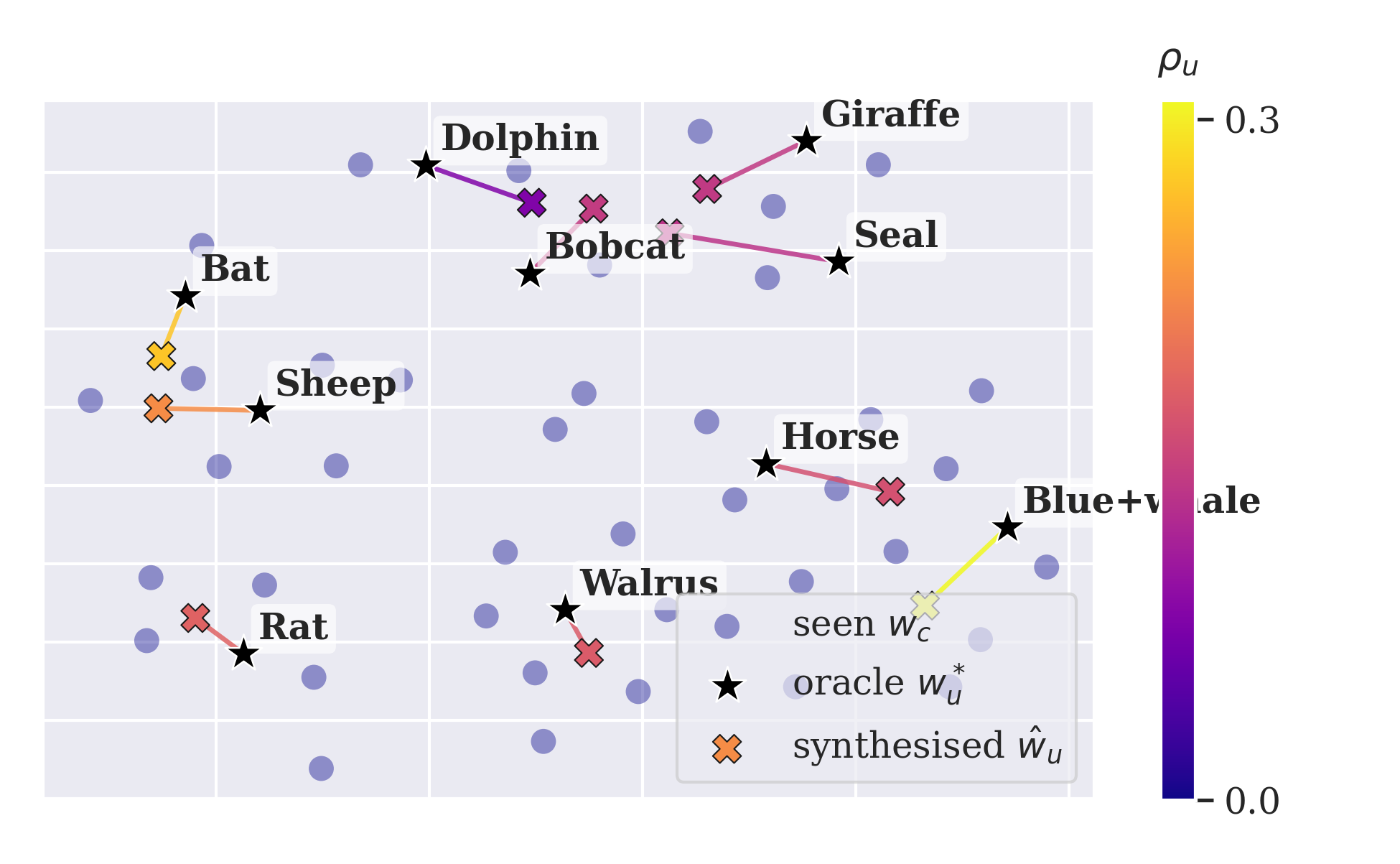}
    \caption{Geometry of $\hat{w}_u$ vs.\ the true $w_u^*$ in weight space (AWA2, t-SNE). Stars: $w_u^*$ from a classifier retrained per held-out class. Crosses: synthesised $\hat{w}_u$. Segment colour: $\rho_u$. Low-$\rho_u$ classes synthesise close to their true weight; Blue+whale, the highest-$\rho_u$ class, shows the largest gap.}
    \label{fig:geometry}
\end{figure}

The mechanism is visible directly in weight space. Figure~\ref{fig:geometry} projects seen weights, the true $w_u^*$ from a classifier retrained on each held-out class and $\hat{w}_u$ into a shared t-SNE embedding for the AWA2 unseen classes, with the $\hat{w}_u$-to-$w_u^*$ segment coloured by $\rho_u$. Near-zero-$\rho_u$ classes (Dolphin, Bobcat, Giraffe, Seal) synthesise almost on top of their true weight, while Blue+whale, the $\rho_u$ outlier, shows the largest displacement.

\begin{figure}[b]
    \centering
    \begin{subfigure}{.4\textwidth}
        \centering
        \includegraphics[width=\linewidth]{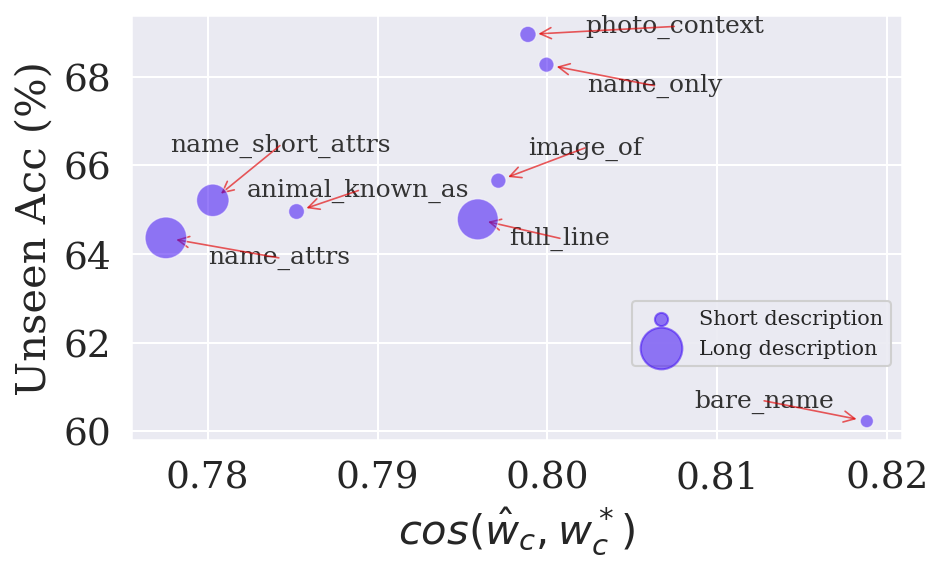}
        \caption{}
        \label{fig:desc_ablation_a}
    \end{subfigure}
    \begin{subfigure}{.4\textwidth}
        \centering
        \includegraphics[width=\linewidth]{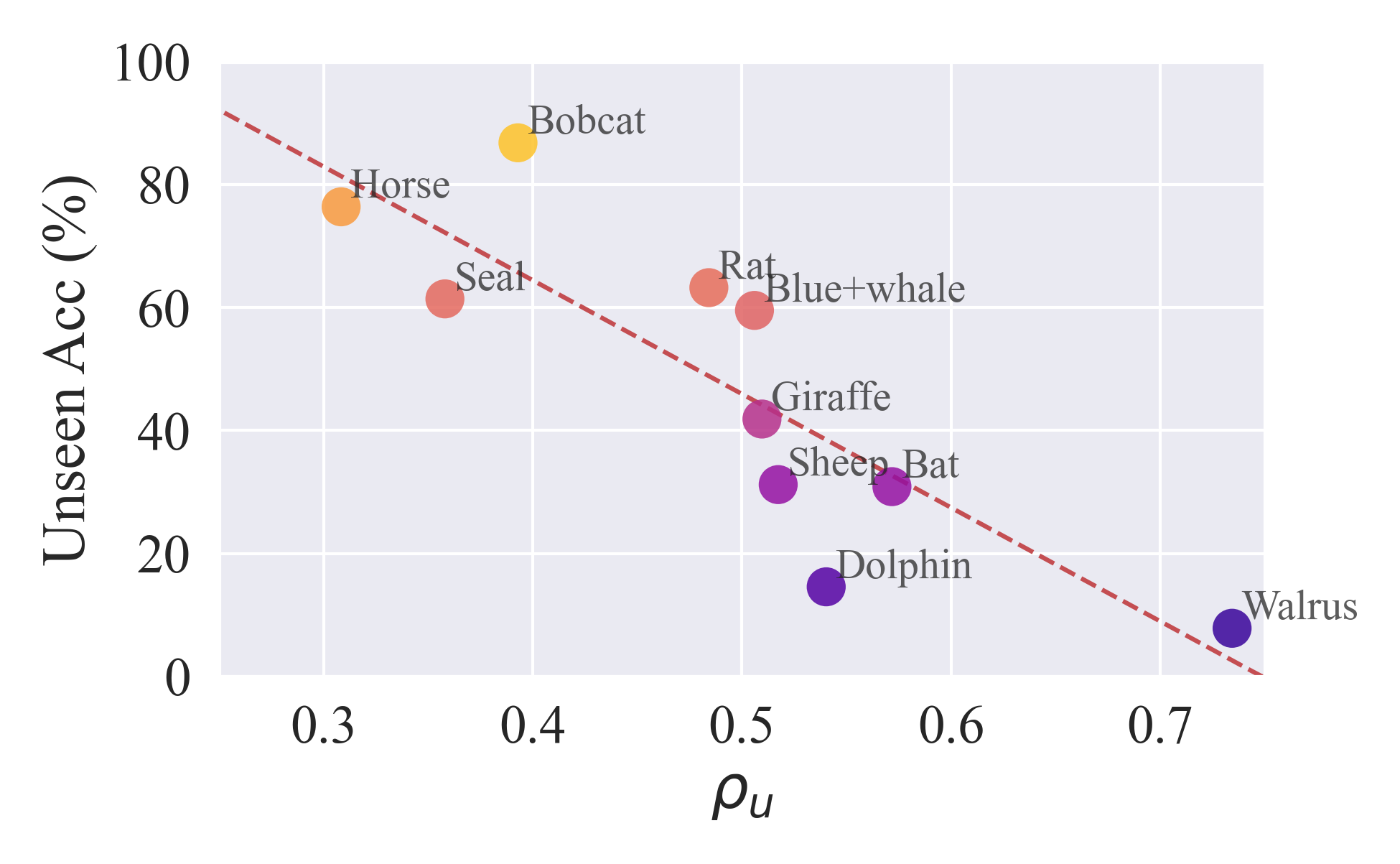}
        \caption{}
        \label{fig:desc_ablation_b}
    \end{subfigure}
\caption{Effect of unseen-class description on $\rho_u$ and synthesis quality, on AWA2. (\subref{fig:desc_ablation_a})~Accuracy vs.\ bridge quality across description templates (listed in the supplementary material). (\subref{fig:desc_ablation_b})~Per-class accuracy vs.\ $\rho_u$.}
\end{figure}

\paragraph{A property of the problem, not of CAST.} The blind-spot argument is not specific to Ridge. Any linear estimator $\hat{P}=f(W,Z)$ shares the same null space $\operatorname{col}(Z)^\perp$. Consequently $\rho_u$ is a lower bound on the extrapolation difficulty. This offers two practical uses. First, as a \textit{per-description screen} where $\rho_u^{(k)} = \|(I-ZZ^\dagger)g(a_u^{(k)})\|$ indicates descriptions that will lead to a poor synthesis regardless of semantic precision. Second, as a \textit{benchmark diagnostic measure} by averaging over all unseen classes,
\begin{equation}
    \label{eq:rho_bar}
    \bar{\rho} = \frac{1}{|\mathcal{Y}^u|}\sum_{u \in \mathcal{Y}^u} \rho_u \in [0,1]
\end{equation}
we characterises how much of a benchmark's unseen semantic space lies outside the seen-class span, independently of method.

Both uses hold up empirically (Figure~\ref{fig:desc_ablation_a}--\ref{fig:desc_ablation_b}). Since CLIP maps different textual descriptions of the same concept to different embedding points, a description determines where $z_u$ lands relative to the fixed span $\operatorname{col}(Z_\text{seen})$. Figure~\ref{fig:desc_ablation_a} shows this with a fixed bridge, measured by the class average reconstruction-fidelity score. This estimates how well-conditioned the text-to-weight bridge is for a given description style. Looping through eight descriptions (listed in the supplementary material) while holding $Z_\text{seen}$ fixed produces up to ${\sim}8$ points of accuracy variation, from differences in both $\rho_u$ and the discriminativeness of $z_u^\parallel$. Fixing the descriptions instead, $\rho_u$ characterises how hard each unseen class is for \emph{any} linear method. Figure~\ref{fig:desc_ablation_b} confirms this, with an inverse relationship between accuracy and $\rho_u$.

\section{Experiments}
We evaluate CAST on four standard ZSL benchmarks , the AWA2~\cite{xian2018zero}, the CUB~\cite{welinder2010caltech}, APY~\cite{farhadi2009describing}, and FLO~\cite{nilsback2008automated}. Using their conventional proposed splits~\cite{xian2018zero} under both the conventional ZSL setting and the Generalised ZSL (GZSL) setting. Under GZSL performance is measured by the harmonic mean $H = \frac{2 \times s \times u}{s + u}$ of seen ($s$) and unseen ($u$) accuracies. All pre-trained classifiers are from PyTorch's Computer Vision library~\cite{torchvision2016}; CLIP models are from the OpenAI library~\cite{radford2021learning}.
\begin{table}[]
\centering
\small
\renewcommand{\arraystretch}{0.85}
\begin{tabular}{@{}rc|llll@{}}
\toprule
\multicolumn{1}{l}{\textit{AWA2}}        &          & \multicolumn{4}{c}{\textbf{CLIP}}                                                                                                     \\ \cmidrule(l){2-6} 
\multicolumn{1}{r|}{\textbf{Classifier}} & Baseline & \multicolumn{1}{c}{ViT-b32}     & \multicolumn{1}{c}{ViT-b16}     & \multicolumn{1}{c}{RN50}        & \multicolumn{1}{c}{RN101}       \\ \midrule
\multicolumn{1}{r|}{AlexNet}             & 92.68    & 63.5 (92.4)                     & 64.0 (92.1)                     & 60.1 (92.6)                     & 68.3 (92.5)                     \\
\multicolumn{1}{r|}{ConvNext B}          & 96.13    & 72.1 (96.3)                     & \textcolor{red}{78.5} (96.3)                     & \textcolor{red}{74.9} (96.2)                     & 75.7 (96.2)                     \\
\multicolumn{1}{r|}{DensNet161}          & 95.49    & 72.0 (95.4)                     & 68.9 (95.3)                     & 64.8 (95.4)                     & 73.2 (95.4)                     \\
\multicolumn{1}{l|}{EfficientNet-v2-M}   & 89.49    & \textcolor{blue}{72.5} (89.4)                    & 71.1 (89.2)                     & 71.5 (89.5)                     & 72.1 (89.4)                     \\
\multicolumn{1}{r|}{GoogLeNet}           & 89.17    & 67.9 (89.2)                     & 68.3 (88.9)                       & 61.5 (89.2)                    & 68.3 (89.2)                             \\
\multicolumn{1}{r|}{MobileNet v2}        & 92.20    & \textcolor{blue}{72.5} (92.1)                     & 68.9 (92.0)                     & 64.9 (92.1)                     & 74.5 (92.2)                     \\
\multicolumn{1}{r|}{ShuffleNet v2 x1}    & 42.98    & 47.8 (41.5)                     & 54.4 (37.3)                     & 39.2 (42.2)                     & 25.4 (40.7)                     \\
\multicolumn{1}{r|}{ResNet 50}           & 94.29    & \multicolumn{1}{c}{\textcolor{red}{76.5} (94.2)} & \multicolumn{1}{c}{72.0 (94.0)} & \multicolumn{1}{c}{\textcolor{red}{74.9} (94.2)} & \multicolumn{1}{c}{\textcolor{blue}{79.4} (94.2)} \\
\multicolumn{1}{r|}{Swin v2 B}           & 95.47    & 69.1 (95.6)                     & \textcolor{blue}{76.7} (95.2)                     & \textcolor{blue}{73.6} (95.6)                     & \textcolor{red}{79.8} (95.4)                     \\
\multicolumn{1}{r|}{ViT-b16}             & 95.41    & \multicolumn{1}{c}{65.9 (95.3)} & \multicolumn{1}{c}{59.7 (95.3)} & \multicolumn{1}{c}{63.4 (95.3)} & \multicolumn{1}{c}{68.5 (95.3)} \\
\multicolumn{1}{r|}{Wide ResNet 50}      & 93.77    & \multicolumn{1}{c}{71.9 (93.7)} & \multicolumn{1}{c}{64.9 (93.6)} & \multicolumn{1}{c}{63.9 (93.7)} & \multicolumn{1}{c}{71.5 (93.7)} \\ \bottomrule
\end{tabular}
\caption{Zero-shot classification performance on unseen classes using our proposed CAST method, in the conventional zero-shot setting. \textcolor{red}{Red} and \textcolor{blue}{blue} values indicates best and second best performance of classifier per CLIP model. Parenthetical values report top-1 accuracy for synthesized seen-class weights, reflecting bridge reconstruction quality.}
\label{tab:pre_trained_awa}
\end{table}

\subsection{Cross-Architecture Generalisation}
CAST's theoretical argument makes no assumption about the specific classifier or CLIP backbone, since the linear bridge follows from near-optimal representations of a shared visual-semantic world. We test this across a grid of $11$ pre-trained classifiers and $4$ CLIP text encoders on AWA2 (Table~\ref{tab:pre_trained_awa}; full CUB grid in the supplementary material). CAST generalises consistently on AWA2 (60--80\% unseen accuracy across most pairings), with seen-class reconstruction closely tracking each classifier's baseline (e.g.\ ResNet50: 94.29 vs.\ 94.2). No single CLIP backbone dominates, so backbone choice is secondary. The AWA2-vs-CUB gap (60--80\% vs.\ 22--36\%) mirrors semantic extrapolation difficulty (Section~\ref{sec:theoretical_analysis}): AWA2 unseen classes share broad attributes well-covered by the seen vocabulary, while CUB's fine-grained distinctions largely are not.
\begin{table}[t]
\centering
\small
\renewcommand{\arraystretch}{0.85}
\begin{tabular}{@{}lccccccccclll@{}}
\toprule
\multicolumn{1}{c}{} & \multicolumn{3}{c}{AWA2} & \multicolumn{3}{c}{CUB} & \multicolumn{3}{c}{APY} & \multicolumn{3}{c}{FLO}
\\ \midrule
\multicolumn{1}{l|}{Method} & S & U & \multicolumn{1}{c|}{H} & S & U & \multicolumn{1}{c|}{H} & S & U & \multicolumn{1}{c|}{H} & S & U & H
\\ \midrule

\multicolumn{1}{l|}{\textit{CLIP Base}} & \textit{41.9} & \textit{61.3} & \multicolumn{1}{c|}{\textit{49.8}} & \textit{35.4} & \textit{43.1} & \multicolumn{1}{c|}{\textit{38.9}} & \textit{21.4} & \textit{11.2} & \multicolumn{1}{c|}{\textit{14.7}} & \textit{50.6} & \textit{37.3} & \textit{42.9} \\

\specialrule{1.2pt}{1pt}{1pt}

\multicolumn{1}{l|}{ConSE~\cite{norouzi2013zero}} & 96.1 & 3.0  & \multicolumn{1}{c|}{5.7} & {\color[HTML]{FE0000} 88.0} & 0.5  & \multicolumn{1}{c|}{0.9} & {\color[HTML]{FE0000} 90.5} & 2.8  & \multicolumn{1}{c|}{5.5}  & 98.5  & 0.3 & 0.7 \\

\multicolumn{1}{l|}{COSTA~\cite{mensink2014costa}} & {\color[HTML]{FE0000} 96.1} & 0.0 & \multicolumn{1}{c|}{0.0} & {\color[HTML]{6434FC} 87.6} & 0.0 & \multicolumn{1}{c|}{0.0} & {\color[HTML]{6434FC} 90.2} & 0.0 & \multicolumn{1}{c|}{0.0} & {\color[HTML]{FE0000} 98.7} & 0.0 & 0.0 \\

\multicolumn{1}{l|}{SMO~\cite{xu2022vgse}}   & \multicolumn{1}{l}{92.4} & \multicolumn{1}{l}{31.8} & \multicolumn{1}{l|}{47.3} & \multicolumn{1}{l}{52.3} & \multicolumn{1}{l}{{\color[HTML]{6434FC} 39.2}} & \multicolumn{1}{l|}{44.8} & \multicolumn{1}{l}{80.9} & \multicolumn{1}{l}{{\color[HTML]{6434FC} 17.7}} & \multicolumn{1}{l|}{{\color[HTML]{6434FC} 29.0}} & 94.7 & 18.8 & 31.4 \\

\multicolumn{1}{l|}{WAvg~\cite{xu2022vgse}}  & \multicolumn{1}{l}{92.4} & \multicolumn{1}{l}{5.5} & \multicolumn{1}{l|}{10.4} & \multicolumn{1}{l}{52.3} & \multicolumn{1}{l}{1.9} & \multicolumn{1}{l|}{3.7} & \multicolumn{1}{l}{77.9} & \multicolumn{1}{l}{7.4} & \multicolumn{1}{l|}{13.5} & {\color[HTML]{3531FF} 94.5} & 7.4 & 7.1 \\

\multicolumn{1}{l|}{ICIS~\cite{christensen2023image}}  & {\color[HTML]{6434FC} 93.3} & {\color[HTML]{3531FF} 35.6} & \multicolumn{1}{c|}{{\color[HTML]{6434FC} 51.6}} & 73.7 & {\color[HTML]{FE0000} 45.8} & \multicolumn{1}{c|}{{\color[HTML]{FE0000} 56.5}} & 87.9 & 7.6 & \multicolumn{1}{c|}{14.1} & 97.6  & {\color[HTML]{3531FF} 23.4} & {\color[HTML]{6434FC} 37.8} \\

\specialrule{0.5pt}{1pt}{1pt}

\multicolumn{1}{l|}{Ours}  & 89.6 & {\color[HTML]{FE0000} 46.2} & \multicolumn{1}{c|}{{\color[HTML]{FE0000} 60.9}} & 85.2 & 32.4 & \multicolumn{1}{c|}{{\color[HTML]{6434FC} 46.9}} &  77.9 & {\color[HTML]{FE0000} 18.2} & \multicolumn{1}{c|}{{\color[HTML]{FE0000} 29.6}} & 85.4 & {\color[HTML]{FE0000} 51.3} & {\color[HTML]{FE0000} 64.1} \\ \bottomrule
\end{tabular}
\vspace{.05cm}
\caption{Generalized zero-shot learning results. The compared methods are image-free only, i.e.\ they are trained through gradient descent. Our model is the only training-free approach. CLIP linear prope is shown for reference. Best and second-best results per dataset are shown in \textcolor{red}{red} and \textcolor{blue}{blue}, respectively.}
\label{tab:main_results}
\end{table}

\subsection{Comparison with Baseline Methods}
Table~\ref{tab:main_results} reports GZSL performance against image-free baselines that, unlike CAST, require gradient-based training to learn the visual-semantic mapping. CAST leads on AWA2 ($H{=}60.9$) and FLO ($H{=}64.1$), ahead of the strongest baseline ICIS \cite{christensen2023image} by 9.3 and 23.7 points, driven mainly by unseen-class accuracy ($U{=}46.2$ vs.\ $35.6$ on AWA2; $U{=}51.3$ vs.\ $23.4$ on FLO). ConSE \cite{norouzi2013zero} and COSTA \cite{mensink2014costa} reach high seen accuracy but near-zero unseen accuracy, collapsing to seen-class predictions. On CUB, where fine-grained distinctions impose an extrapolation penalty a gradient-based encoder can partly capture but a linear bridge cannot, ICIS leads ($H{=}56.5$ vs.\ CAST's $46.9$); APY is similar, with CAST ($H{=}29.6$) comparable to SMO \cite{xu2022vgse} ($29.0$), ahead of ICIS ($14.1$).

\begin{figure}[!h]
    \centering
    \begin{subfigure}{.4\textwidth}
        \centering
        \includegraphics[width=\linewidth]{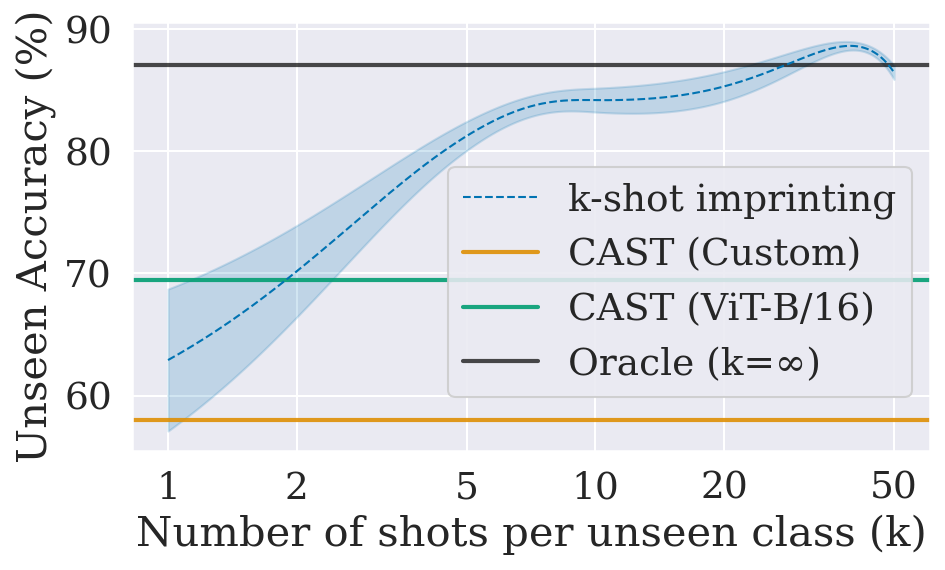}
        \caption{}
        \label{fig:k_shot_curve}
    \end{subfigure}
    \begin{subfigure}{.4\textwidth}
        \centering
        \includegraphics[width=\linewidth]{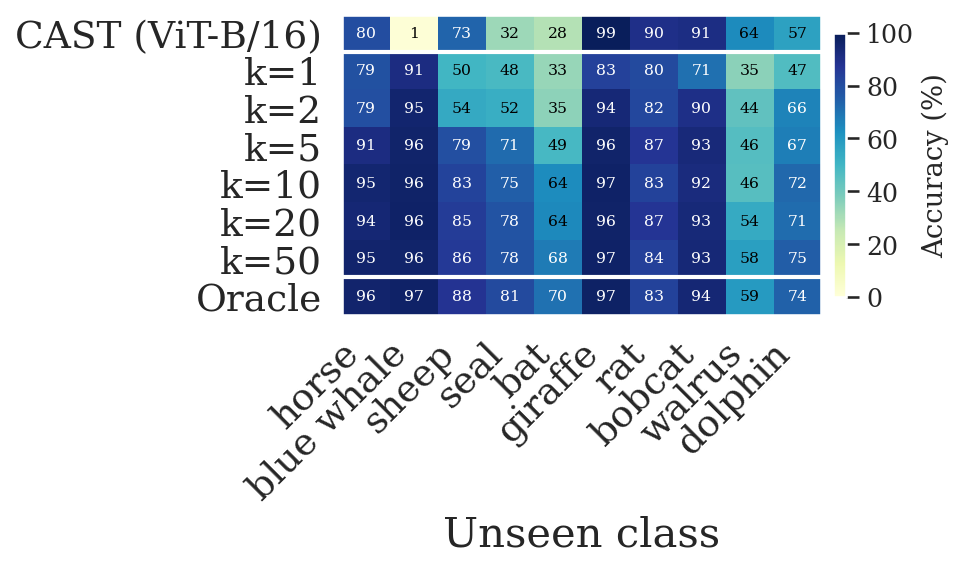}
        \caption{}
        \label{fig:k_shot_heatmap}
    \end{subfigure}
\caption{Imprinting oracle experiment on AWA2. (\subref{fig:k_shot_curve})~Accuracy vs.\ $k$ for $k$-shot imprinting against CAST and the oracle ($k=\infty$). (\subref{fig:k_shot_heatmap})~Per-class breakdown, tracking $\rho_u$ (Section~\ref{sec:semantic_extrapolation_residual}).}
\end{figure}

\subsection{Comparison to $k$-shot Imprinting}
$k$-shot imprinting generates class weights by averaging penultimate-layer embeddings of $k$ labeled examples~\cite{qi2018low}; CAST is represents the $k=0$ limit. Figure~\ref{fig:k_shot_curve} traces accuracy from CAST through $k=1,\ldots$ to the oracle on AWA2.

CAST (ViT-B/16) reaches $\approx$70\% unseen accuracy with zero labeled images, with a semantic data equivalence of $k^*\approx2$. While $k$-shot rises steeply to $k\approx5$ before levelling off, where the bottleneck shifts from data scarcity to representational capacity. The semantic encoder matters, a 2-layer CAST trained on this dataset's own attributes trails ViT-B/16 by $\approx$11 points, confirming bridge quality drives performance. The per-class breakdown (Figure~\ref{fig:k_shot_heatmap}) reveals the non uniformly description quality. CAST matches or exceeds the oracle on some classes (giraffe, rat, walrus) but lags on others (blue whale, bat), tracking $\rho_u$.

\section{Conclusion}
We introduced CAST, an image-free, training-free method for extending pre-trained classifiers to unseen categories via a closed-form Ridge projection. Neither sample images nor iterative training or optimisation are required. A mapping from CLIP text embeddings to a classifier weight space is geometrically account for by unifying Neural Collapse, CLIP cross-modal alignment, and the Linear Representation Hypothesis. A finite-sample bound whose extrapolation term, the \emph{semantic extrapolation residual} $\rho_u$, is computable from text embeddings alone, independent of classifier or synthesis method. Across four ZSL benchmarks and eleven architectures, CAST matches or exceeds gradient-trained image-free baselines with no target-domain images or iterative optimization, and $\rho_u$ predicts where synthesis succeeds or fails, per class and benchmark.

The same analysis marks CAST's boundary. As the bridge is linear, it inherits the null space of the seen-class span, so fine-grained benchmarks (CUB) remain harder than broad-coverage ones (AWA2). Closing this gap while keeping CAST closed-form and training-free, via richer maps or $\rho_u$-guided curation, is future work.

\newpage
\appendix
\section{Supplementary Material}
\label{sec:appendix}
This supplementary material provides the two derivations from Sections~\ref{sec:ridge_objective} and~\ref{sec:theoretical_analysis} of the main paper -- the shrinkage form of the Ridge estimator (Eq.~\eqref{eq:ridge_svd}) and the full error bound behind Proposition~\ref{prop:bound} -- together with additional experimental results and configuration details referenced there.

\subsection{From the Ridge Objective to the Shrinkage Form}
\label{app:shrinkage}
Differentiating the Ridge objective (Eq.~\eqref{eq:ridge}) with respect to $P$ and setting the gradient to zero,
\begin{equation}
    2(PZ - W)Z^\top + 2\lambda P = 0
    \;\;\Longrightarrow\;\;
    P(ZZ^\top + \lambda I) = WZ^\top,
\end{equation}
which rearranges directly to the closed form $\hat{P} = WZ^\top(ZZ^\top+\lambda I)^{-1}$ (Eq.~\eqref{eq:ridge_solution}).

To see how $\hat P$ treats each semantic direction, write the thin SVD $Z = U\Sigma V^\top$, with $U \in \mathbb{R}^{m \times r}$, $V \in \mathbb{R}^{|\mathcal{Y}^s| \times r}$ orthonormal ($U^\top U = V^\top V = I_r$) and $\Sigma = \operatorname{diag}(\sigma_1, \dots, \sigma_r)$. Since $ZZ^\top = U\Sigma^2 U^\top$ acts as $\Sigma^2$ on $\operatorname{col}(U)$ and as $0$ on its orthogonal complement,
\begin{equation}
    (ZZ^\top + \lambda I)^{-1} = U(\Sigma^2+\lambda I)^{-1}U^\top + \tfrac{1}{\lambda}(I - UU^\top).
\end{equation}
Left-multiplying by $Z^\top = V\Sigma U^\top$ and using $U^\top U = I_r$ kills the second term entirely ($U^\top(I-UU^\top) = 0$), leaving
\begin{equation}
    Z^\top(ZZ^\top+\lambda I)^{-1} = V\Sigma(\Sigma^2+\lambda I)^{-1}U^\top = V\operatorname{diag}\!\left(\frac{\sigma_k}{\sigma_k^2+\lambda}\right)U^\top.
\end{equation}
Substituting into $\hat{P} = WZ^\top(ZZ^\top+\lambda I)^{-1}$ gives exactly the shrinkage form of Eq.~\eqref{eq:ridge_svd}: $\hat{P}_\lambda = WV\operatorname{diag}\!\left(\sigma_k/(\sigma_k^2+\lambda)\right)U^\top$. Each seen-class semantic direction $u_k$ is therefore weighted by $\sigma_k/(\sigma_k^2+\lambda)$ rather than the raw $1/\sigma_k$ used by the unregularised solution. This is exactly the shrinkage behaviour discussed in Section~\ref{sec:ridge_objective}.

\subsection{Full Derivation of the CAST Error Bound}
\label{app:bound}
Proposition~\ref{prop:bound} states $\|\hat{w}_u - w_u^*\| \leq O(\eta + \delta + \|P^*\|_{\operatorname{op}}\rho_u)$. We now derive this bound in full, including the constant factor and one additional term that the $O(\cdot)$ in the main text absorbs.

\paragraph{Two sources of error.} Inserting $P^*z_u$ as a pivot splits the total error into two independent pieces:
\begin{equation}
    \|\hat{w}_u - w_u^*\| = \|\hat{P}z_u - w_u^*\|
    \leq \underbrace{\|\hat{P}z_u - P^*z_u\|}_{E_1 \text{ (estimation error)}}
    + \underbrace{\|P^*z_u - w_u^*\|}_{E_2 \text{ (model error)}}.
\end{equation}
$E_2$ is the error we would still pay even with perfect knowledge of the ideal map $P^*$; $E_1$ is the extra cost of only having the Ridge estimate $\hat{P}$ computed from seen classes.

\paragraph{Model error $E_2$.} Inserting $\hat{\mu}_u$ as a second pivot and applying A3 then A1 (as in Section~\ref{sec:theoretical_analysis}),
\begin{equation}
    E_2 \leq \|P^*z_u - \hat{\mu}_u\| + \|\hat{\mu}_u - w_u^*\|
    \leq \big(\delta + \|P^*\|_{\operatorname{op}}\rho_u\big) + \eta.
\end{equation}

\paragraph{Estimation error $E_1$.} Using the span decomposition $z_u = z_u^\parallel + z_u^\perp$ and the blind-spot fact $\hat{P}z_u^\perp = 0$ from Section~\ref{sec:semantic_extrapolation_residual},
\begin{equation}
    E_1 \leq \underbrace{\|(\hat{P}-P^*)z_u^\parallel\|}_{\text{in-span disagreement}} + \underbrace{\|P^*z_u^\perp\|}_{\leq\, \|P^*\|_{\operatorname{op}}\rho_u}.
\end{equation}
The in-span term measures how much $\hat{P}$ and $P^*$ disagree on directions both can see. Writing $Z=U\Sigma V^\top$ as in Section~\ref{app:shrinkage}, this disagreement is controlled by (i) how well-spread the seen-class embeddings are, via $\sigma_{\min}(Z)$, and (ii) $\xi = \|W(I-VV^\top)\|_F$, which measures whether classes that look similar in CLIP text space are also similarly related in weight space, i.e. how tightly representation equivalence holds. A standard Ridge bias-variance argument gives
\begin{equation}
    \|(\hat{P}-P^*)z_u^\parallel\| \leq
    \frac{\big(\xi + \sqrt{|\mathcal{Y}^s|}\,\delta\big)\|z_u\|}{\sigma_{\min}(Z)}
    \;+\; \frac{\lambda\|W\|\,\|z_u\|}{\sigma_{\min}(Z)\big(\sigma_{\min}^2(Z)+\lambda\big)},
\end{equation}
where the second term is the Ridge shrinkage bias, negligible whenever $\lambda \ll \sigma_{\min}^2(Z)$.

\paragraph{Combining.} Summing $E_1$ and $E_2$,
\begin{equation}
\label{eq:bound_full}
    \|\hat{w}_u - w_u^*\| \leq
    \underbrace{\eta + \delta}_{\text{intrinsic}}
    + \underbrace{2\|P^*\|_{\operatorname{op}}\rho_u}_{\text{extrapolation}}
    + \underbrace{\frac{(\xi + \sqrt{|\mathcal{Y}^s|}\,\delta)\|z_u\|}{\sigma_{\min}(Z)}}_{\text{conditioning}}.
\end{equation}
This is the bound in full: the extrapolation penalty carries a factor of $2$ because $z_u^\perp$ costs twice -- once because $\hat{P}$ cannot see it at all, and once because even the ideal $P^*$ only bounds its own error there by $\|P^*\|_{\operatorname{op}}\rho_u$. The third conditioning term reduces to $O(\delta)$ in practice as CLIP's contrastive training spreads embeddings roughly uniformly (the ETF geometry from NC and CLIP alignment ensures this). The main text does not carry this conditioning term explicitly.

\paragraph{Recovering Proposition~\ref{prop:bound}.} The conditioning term vanishes under one further, mild assumption: \textbf{(A4)} the seen-class embeddings are not near-collinear, $\sigma_{\min}(Z) = \Omega(\sqrt{|\mathcal{Y}^s|})$, and $\xi \approx 0$. Both hold in practice. CLIP's contrastive training spreads text embeddings roughly uniformly across the semantic space (a consequence of the same near-optimality behind A2 and A3), so $\sigma_{\min}(Z)$ collapsing would require many seen classes to be near-synonymous, an easily detected labelling pathology rather than a property of realistic benchmarks. Under A4, the conditioning term in Eq.~\eqref{eq:bound_full} is $O(\delta)$, and the constant factor on the extrapolation term is absorbed into the asymptotic notation, giving exactly
\begin{equation}
    \|\hat{w}_u - w_u^*\| \leq O\big(\eta + \delta + \|P^*\|_{\operatorname{op}}\rho_u\big),
\end{equation}
i.e.\ Proposition~\ref{prop:bound}.

\subsection{Additional Cross-Architecture Results}
\label{app:extra_results}
Table~\ref{tab:pretrained_cub} gives the full CUB cross-architecture grid, evaluating the same $11$ pre-trained classifiers against $4$ CLIP text encoders summarised in Section~6.1 of the main paper and reported for AWA2 in Table~\ref{tab:pre_trained_awa}; both use the same format, with parenthetical values reporting the seen-class reconstruction accuracy.
\begin{table}[]
\centering
\small
\begin{tabular}{@{}rccccc@{}}
\toprule
\multicolumn{1}{l}{\textit{CUB}}         & \multicolumn{1}{c|}{}         & \multicolumn{4}{c}{\textbf{CLIP}}                                                                 \\ \midrule
\multicolumn{1}{r|}{\textbf{Classifier}} & \multicolumn{1}{c|}{Baseline} & RN50                   & RN101                  & ViT-B-32               & ViT-B-16               \\ \midrule
AlexNet                                  & 78.34                           & 30.1 (76.3)          & 30.9 (74.9)          & 30.74 (74.2)          & 30.1 (74.6)          \\
ConvNext                                 & 54.99                           & 30.0 (54.1)          & 29.2 (54.1)          & 32.9 (51.5)          & 29.4 (54.0)          \\
DenseNet                                 & 67.60                           & 33.3 (66.9)          & 36.1 (66.4)          & 36.2 (66.0)          & 35.5 (65.8)          \\
EfficientNet-v2-M                        & 43.20                           & 22.0 (43.3)          & 23.1 (42.9)          & 24.3 (42.5)          & 21.7 (43.1)          \\
GoogLeNet                                & 44.66           & 25.2 (43.5) & 25.6 (42.9) & 24.4 (42.5) & 22.4 (42.6) \\
MobileNet v2                                & 58.80                           & 30.7 (58.7)          & 31.9 (58.2)          & 30.4 (57.5)          & 30.8 (57.9)          \\
ResNet 50                                & 63.53                           & 32.1 (62.8)          & 32.8 (62.9)          & 35.5 (61.5)          & 31.2 (61.8)          \\
ShuffleNet v2 x1                    & 17.91                           & 5.6 (15.6)           & 8.7 (12.1)           & 7.3 (17.0)           & 6.0 (11.2)           \\
Swin v2 B                              & 49.45                           & 29.6 (49.1)          & 28.7 (48.2)          & 30.1 (47.8)          & 27.2 (48.3)          \\
ViT-b16                                 & 56.52                           & 28.0 (56.5)          & 28.9 (55.3)          & 34.8 (54.9)          & 29.6 (56.7)          \\
Wide Resnet 50                              & 55.37                           & 22.2 (55.4)          & 25.9 (55.1)          & 25.8 (54.4)          & 24.1 (53.7)          \\ \bottomrule
\end{tabular}
\vspace{.2cm}
\caption{Zero-shot classification performance on unseen CUB classes, in the same format as Table~\ref{tab:pre_trained_awa}. The consistently lower absolute accuracy relative to AWA2 reflects CUB's higher semantic extrapolation difficulty $\bar{\rho}$ (Section~\ref{sec:semantic_extrapolation_residual}), not a failure specific to any classifier or CLIP backbone.}
\label{tab:pretrained_cub}
\end{table}

\subsection{Description Templates}
\label{app:extra_validation}
Table~\ref{tab:text_description} lists the eight description templates used in the description-ablation study of Section~\ref{sec:semantic_extrapolation_residual} of the main paper (Figure~\ref{fig:desc_ablation_a}).
\begin{table}
    \centering
    \begin{adjustbox}{max width=\linewidth}
    \begin{tabular}{ll|l}
    \toprule
     & \textbf{Variant} & \textbf{Example description} \\
    \midrule
    1 & \texttt{name\_only}        & ``a photo of a antelope'' \\
    2 & \texttt{image\_of}         & ``an image of an antelope'' \\
    3 & \texttt{photo\_context}    & ``a wildlife photo of an antelope'' \\
    4 & \texttt{bare\_name}        & ``antelope'' \\
    5 & \texttt{full\_line}        & ``antelope: skin/coat: furry, toughskin; size: big; body features: lean, hooves, \ldots'' \\
    6 & \texttt{attrs\_only}       & ``skin/coat: furry, toughskin; size: big; body features: lean, hooves, \ldots'' \\
    7 & \texttt{name\_attrs}       & ``a photo of an antelope. skin/coat: furry, toughskin; size: big; \ldots'' \\
    8 & \texttt{animal\_known\_as} & ``an animal known as antelope'' \\
    \bottomrule
    \end{tabular}
    \end{adjustbox}
    \caption{Textual description variants evaluated for encoding unseen classes via CLIP. Each variant represents a different design choice for how semantic information is presented.}
    \label{tab:text_description}
\end{table}


\newpage

\begin{thebibliography}{10}
\providecommand{\url}[1]{\texttt{#1}}
\providecommand{\urlprefix}{URL }
\providecommand{\doi}[1]{https://doi.org/#1}

\bibitem{akata2013label}
Akata, Z., Perronnin, F., Harchaoui, Z., Schmid, C.: Label-embedding for
  attribute-based classification. In: Proceedings of the IEEE conference on
  computer vision and pattern recognition. pp. 819--826 (2013)

\bibitem{akata2015evaluation}
Akata, Z., Reed, S., Walter, D., Lee, H., Schiele, B.: Evaluation of output
  embeddings for fine-grained image classification. In: Proceedings of the IEEE
  conference on computer vision and pattern recognition. pp. 2927--2936 (2015)

\bibitem{bucher2016improving}
Bucher, M., Herbin, S., Jurie, F.: Improving semantic embedding consistency by
  metric learning for zero-shot classiffication. In: European Conference on
  Computer Vision. pp. 730--746. Springer (2016)

\bibitem{bucher2017generating}
Bucher, M., Herbin, S., Jurie, F.: Generating visual representations for
  zero-shot classification. In: Proceedings of the IEEE International
  Conference on Computer Vision Workshops. pp. 2666--2673 (2017)

\bibitem{chen2023deconstructed}
Chen, D., Shen, Y., Zhang, H., Torr, P.H.: Deconstructed generation-based
  zero-shot model. In: Proceedings of the AAAI Conference on Artificial
  Intelligence. vol.~37, pp. 295--303 (2023)

\bibitem{chen2021knowledge}
Chen, J., Geng, Y., Chen, Z., Horrocks, I., Pan, J.Z., Chen, H.:
  Knowledge-aware zero-shot learning: Survey and perspective. arXiv preprint
  arXiv:2103.00070  (2021)

\bibitem{chen2021review}
Chen, L., Li, S., Bai, Q., Yang, J., Jiang, S., Miao, Y.: Review of image
  classification algorithms based on convolutional neural networks. Remote
  Sensing  \textbf{13}(22), ~4712 (2021)

\bibitem{chen2024understanding}
Chen, Z., Deng, Y., Li, Y., Gu, Q.: Understanding transferable representation
  learning and zero-shot transfer in clip. In: International Conference on
  Learning Representations. vol.~2024, pp. 55415--55444 (2024)

\bibitem{christensen2023image}
Christensen, A., Mancini, M., Koepke, A., Winther, O., Akata, Z.: Image-free
  classifier injection for zero-shot classification. In: Proceedings of the
  IEEE/CVF international conference on computer vision. pp. 19072--19081 (2023)

\bibitem{ciresan2011flexible}
Ciresan, D.C., Meier, U., Masci, J., Maria~Gambardella, L., Schmidhuber, J.:
  Flexible, high performance convolutional neural networks for image
  classification. In: IJCAI proceedings-international joint conference on
  artificial intelligence. vol.~22, p.~1237. Barcelona, Spain: (2011)

\bibitem{dinu2014improving}
Dinu, G., Lazaridou, A., Baroni, M.: Improving zero-shot learning by mitigating
  the hubness problem. arXiv preprint arXiv:1412.6568  (2014)

\bibitem{dong2025adapting}
Dong, H., Sheng, L., Liang, J., He, R., Chatzi, E., Fink, O.: Adapting
  vision-language models without labels: A comprehensive survey. arXiv preprint
  arXiv:2508.05547  (2025)

\bibitem{elhoseiny2021cizsl}
Elhoseiny, M., Yi, K., Elfeki, M.: Cizsl++: Creativity inspired generative
  zero-shot learning. arXiv preprint arXiv:2101.00173  (2021)

\bibitem{farhadi2009describing}
Farhadi, A., Endres, I., Hoiem, D., Forsyth, D.: Describing objects by their
  attributes. In: IEEE Conference on Computer Vision and Pattern Recognition.
  pp. 1778--1785 (2009)

\bibitem{gao2021clip}
Gao, P., Geng, S., Zhang, R., Ma, T., Fang, R., Zhang, Y., Li, H., Qiao, Y.:
  Clip-adapter: Better vision-language models with feature adapters.
  International Journal of Computer Vision  \textbf{132},  581--595 (2024)

\bibitem{hentschel2022clip}
Hentschel, S., Kobs, K., Hotho, A.: Clip knows image aesthetics. Frontiers in
  Artificial Intelligence  \textbf{5},  976235 (2022)

\bibitem{hild2006feature}
Hild, K.E., Erdogmus, D., Torkkola, K., Principe, J.C.: Feature extraction
  using information-theoretic learning. IEEE Transactions on Pattern Analysis
  and Machine Intelligence  \textbf{28}(9),  1385--1392 (2006)

\bibitem{jiang2018learning}
Jiang, H., Wang, R., Shan, S., Chen, X.: Learning class prototypes via
  structure alignment for zero-shot recognition. In: Proceedings of the
  European conference on computer vision (ECCV). pp. 118--134 (2018)

\bibitem{kodirov2017semantic}
Kodirov, E., Xiang, T., Gong, S.: Semantic autoencoder for zero-shot learning.
  In: Proceedings of the IEEE Conference on Computer Vision and Pattern
  Recognition. pp. 4447--4456 (2017)

\bibitem{lenc2015understanding}
Lenc, K., Vedaldi, A.: Understanding image representations by measuring their
  equivariance and equivalence. In: Proceedings of the IEEE conference on
  computer vision and pattern recognition. pp. 991--999 (2015)

\bibitem{li2019alleviating}
Li, J., Jing, M., Lu, K., Zhu, L., Yang, Y., Huang, Z.: Alleviating feature
  confusion for generative zero-shot learning. In: Proceedings of the 27th ACM
  international conference on multimedia. pp. 1587--1595 (2019)

\bibitem{li2017zero}
Li, Y., Wang, D., Hu, H., Lin, Y., Zhuang, Y.: Zero-shot recognition using dual
  visual-semantic mapping paths. In: Proceedings of the IEEE conference on
  computer vision and pattern recognition. pp. 3279--3287 (2017)

\bibitem{lv2023learning}
Lv, F., Zhang, J., Yang, G., Feng, L., Yu, Y., Duan, L.: Learning cross-domain
  semantic-visual relationships for transductive zero-shot learning. Pattern
  Recognition  \textbf{141},  109591 (2023)

\bibitem{lyu2025expanding}
Lyu, D., Wang, X., Ban, T., Chen, L., Zhou, X., Chen, H.: Expanding the
  category of classifiers with {LLM} supervision. In: Proceedings of the
  Thirty-Fourth International Joint Conference on Artificial Intelligence
  (IJCAI). pp. 5905--5913 (2025)

\bibitem{torchvision2016}
maintainers, T., contributors: Torchvision: Pytorch's computer vision library.
  \url{https://github.com/pytorch/vision} (2016)

\bibitem{mensink2014costa}
Mensink, T., Gavves, E., Snoek, C.G.M.: Costa: Co-occurrence statistics for
  zero-shot classification. In: IEEE Conference on Computer Vision and Pattern
  Recognition. pp. 2441--2448 (2014)

\bibitem{nilsback2008automated}
Nilsback, M.E., Zisserman, A.: Automated flower classification over a large
  number of classes. In: Indian Conference on Computer Vision, Graphics \&
  Image Processing. pp. 722--729 (2008)

\bibitem{norouzi2013zero}
Norouzi, M., Mikolov, T., Bengio, S., Singer, Y., Shlens, J., Frome, A.,
  Corrado, G.S., Dean, J.: Zero-shot learning by convex combination of semantic
  embeddings. arXiv preprint arXiv:1312.5650  (2013)

\bibitem{papyan2020prevalence}
Papyan, V., Han, X., Donoho, D.L.: Prevalence of neural collapse during the
  terminal phase of deep learning training. Proceedings of the National Academy
  of Sciences  \textbf{117}(40),  24652--24663 (2020)

\bibitem{park2023linear}
Park, K., Choe, Y.J., Veitch, V.: The linear representation hypothesis and the
  geometry of large language models. arXiv preprint arXiv:2311.03658  (2023)

\bibitem{paul2019semantically}
Paul, A., Krishnan, N.C., Munjal, P.: Semantically aligned bias reducing zero
  shot learning. In: Proceedings of the IEEE/CVF Conference on Computer Vision
  and Pattern Recognition. pp. 7056--7065 (2019)

\bibitem{pittaras2016comparison}
Pittaras, N., Markatopoulou, F., Mezaris, V., Patras, I.: Comparison of
  fine-tuning and extension strategies for deep convolutional neural networks.
  In: International conference on multimedia modeling. pp. 102--114. Springer
  (2016)

\bibitem{qi2018low}
Qi, H., Brown, M., Lowe, D.G.: Low-shot learning with imprinted weights. In:
  Proceedings of the IEEE conference on computer vision and pattern
  recognition. pp. 5822--5830 (2018)

\bibitem{radford2021learning}
Radford, A., Kim, J.W., Hallacy, C., Ramesh, A., Goh, G., Agarwal, S., Sastry,
  G., Askell, A., Mishkin, P., Clark, J., et~al.: Learning transferable visual
  models from natural language supervision. In: International conference on
  machine learning. pp. 8748--8763. PmLR (2021)

\bibitem{rawat2017deep}
Rawat, W., Wang, Z.: Deep convolutional neural networks for image
  classification: A comprehensive review. Neural computation  \textbf{29}(9),
  2352--2449 (2017)

\bibitem{romeraparedes2015embarrassingly}
Romera-Paredes, B., Torr, P.: An embarrassingly simple approach to zero-shot
  learning. In: International Conference on Machine Learning. pp. 2152--2161.
  PMLR (2015)

\bibitem{shigeto2015ridge}
Shigeto, Y., Suzuki, I., Hara, K., Shimbo, M., Matsumoto, Y.: Ridge regression,
  hubness, and zero-shot learning. In: Joint European Conference on Machine
  Learning and Knowledge Discovery in Databases. pp. 135--151. Springer (2015)

\bibitem{song2018transductive}
Song, J., Shen, C., Yang, Y., Liu, Y., Song, M.: Transductive unbiased
  embedding for zero-shot learning. In: Proceedings of the IEEE conference on
  computer vision and pattern recognition. pp. 1024--1033 (2018)

\bibitem{thong2020bias}
Thong, W., Snoek, C.G.: Bias-awareness for zero-shot learning the seen and
  unseen. arXiv preprint arXiv:2008.11185  (2020)

\bibitem{vylomova2016take}
Vylomova, E., Rimell, L., Cohn, T., Baldwin, T.: Take and took, gaggle and
  goose, book and read: Evaluating the utility of vector differences for
  lexical relation learning. In: Proceedings of the 54th annual meeting of the
  association for computational linguistics (volume 1: long papers). pp.
  1671--1682 (2016)

\bibitem{wang2022whose}
Wang, D., Prabhat, S., Sambasivan, N.: Whose ai dream? in search of the
  aspiration in data annotation. In: Proceedings of the 2022 CHI conference on
  human factors in computing systems. pp. 1--16 (2022)

\bibitem{wang2019survey}
Wang, W., Zheng, V.W., Yu, H., Miao, C.: A survey of zero-shot learning:
  Settings, methods, and applications. ACM Transactions on Intelligent Systems
  and Technology (TIST)  \textbf{10}(2),  1--37 (2019)

\bibitem{wang2026counterfactual}
Wang, X., Gao, Y., Rong, C., Chen, L., Lyu, D., Zhou, X., Ban, T., Chen, H.:
  Counterfactual-driven zero-shot classifier expansion. In: Proceedings of the
  AAAI Conference on Artificial Intelligence. vol.~40, pp. 26508--26516 (2026)

\bibitem{welinder2010caltech}
Welinder, P., Branson, S., Mita, T., Wah, C., Schroff, F., Belongie, S.,
  Perona, P.: Caltech-ucsd birds 200. Tech. Rep. CNS-TR-2010-001, California
  Institute of Technology (2010)

\bibitem{xian2018zero}
Xian, Y., Lampert, C.H., Schiele, B., Akata, Z.: Zero-shot learning—a
  comprehensive evaluation of the good, the bad and the ugly. IEEE transactions
  on pattern analysis and machine intelligence  \textbf{41}(9),  2251--2265
  (2018)

\bibitem{xian2019f}
Xian, Y., Sharma, S., Schiele, B., Akata, Z.: f-vaegan-d2: A feature generating
  framework for any-shot learning. In: Proceedings of the IEEE/CVF conference
  on computer vision and pattern recognition. pp. 10275--10284 (2019)

\bibitem{xu2022vgse}
Xu, W., Xian, Y., Wang, J., Schiele, B., Akata, Z.: Vgse: Visually-grounded
  semantic embeddings for zero-shot learning. In: Proceedings of the IEEE/CVF
  Conference on Computer Vision and Pattern Recognition. pp. 9316--9325 (2022)

\bibitem{zhang2024bridging}
Zhang, C., Jin, M., Yu, Q., Xue, H., Gowda, S.N., Jin, X.: Bridging the
  projection gap: Overcoming projection bias through parameterized distance
  learning. In: Proceedings of the Asian Conference on Computer Vision. pp.
  3327--3343 (2024)

\bibitem{zhang2020towards}
Zhang, L., Wang, P., Liu, L., Shen, C., Wei, W., Zhang, Y., Van Den~Hengel, A.:
  Towards effective deep embedding for zero-shot learning. IEEE Transactions on
  Circuits and Systems for Video Technology  \textbf{30}(9),  2843--2852 (2020)

\bibitem{zhou2025revisiting}
Zhou, D.W., Cai, Z.W., Ye, H.J., Zhan, D.C., Liu, Z.: Revisiting
  class-incremental learning with pre-trained models: Generalizability and
  adaptivity are all you need. International Journal of Computer Vision
  \textbf{133}(3),  1012--1032 (2025)

\bibitem{zhou2022conditional}
Zhou, K., Yang, J., Loy, C.C., Liu, Z.: Conditional prompt learning for
  vision-language models. In: Proceedings of the IEEE/CVF conference on
  computer vision and pattern recognition. pp. 16816--16825 (2022)

\bibitem{zhou2022learning}
Zhou, K., Yang, J., Loy, C.C., Liu, Z.: Learning to prompt for vision-language
  models. International journal of computer vision  \textbf{130}(9),
  2337--2348 (2022)

\bibitem{zhu2019semantic}
Zhu, Y., Xie, J., Tang, Z., Peng, X., Elgammal, A.: Semantic-guided
  multi-attention localization for zero-shot learning. Advances in Neural
  Information Processing Systems  \textbf{32} (2019)

\end{thebibliography}
\end{document}